\documentclass[lettersize,journal]{IEEEtran}

\usepackage{booktabs}
\usepackage{amsmath,amsfonts}
\usepackage{algorithmic}
\usepackage{algorithm}
\usepackage{array}
\usepackage[caption=false,font=normalsize,labelfont=sf,textfont=sf]{subfig}
\usepackage{textcomp}
\usepackage{stfloats}
\usepackage{url}
\usepackage{verbatim}
\usepackage{graphicx}
\usepackage{cite}
\begin{document}

\title{TS-VLM: Text-Guided SoftSort Pooling for Vision-Language Models in Multi-View Driving Reasoning}

\author{
Lihong Chen,~Hossein Hassani,~and Soodeh Nikan

\thanks{The authors are with the Department of Electrical and Computer Engineering, Western University, London,
ON N6A 3K7, Canada (e-mail: lchen893@uwo.ca; hhassa52@uwo.ca; snikan@uwo.ca)}}

\maketitle

\begin{abstract}
Vision-Language Models (VLMs) have shown remarkable potential in advancing autonomous driving by leveraging multi-modal fusion in order to enhance scene perception, reasoning, and decision-making. Despite their potential, existing models suffer from computational overhead and inefficient integration of multi-view sensor data that make them impractical for real-time deployment in safety-critical autonomous driving applications. To address these shortcomings, this paper is devoted to designing a lightweight VLM called TS-VLM, which incorporates a novel Text-Guided SoftSort Pooling (TGSSP) module. By resorting to semantics of the input queries, TGSSP ranks and fuses visual features from multiple views, enabling dynamic and query-aware multi-view aggregation without reliance on costly attention mechanisms. This design ensures the query-adaptive prioritization of semantically related views, which leads to improved contextual accuracy in multi-view reasoning for autonomous driving. Extensive evaluations on the DriveLM benchmark demonstrate that, on the one hand, TS-VLM outperforms state-of-the-art models with a BLEU-4 score of 56.82, METEOR of 41.91, ROUGE-L of 74.64, and CIDEr of 3.39. On the other hand, TS-VLM reduces computational cost by up to 90\%, where the smallest version contains only 20.1 million parameters, making it more practical for real-time deployment in autonomous vehicles. 
\end{abstract}

\begin{IEEEkeywords}
Autonomous Driving, Question Answering, Scene Understanding, Vision-Language Model, Vision-Language Reasoning, Visual Perception.
\end{IEEEkeywords}

\section{Introduction}
\IEEEpubidadjcol
The modular pipeline in conventional autonomous driving systems wires together separate perception, prediction, and planning modules. While this decomposition simplifies the development process, it can lead to brittle performance and cascading errors. For instance, rule-based planning modules struggle to handle the vast diversity of real-world driving scenarios. Each component is typically specialized and extensively tuned for specific conditions, making the whole system data-hungry and prone to breakdown under distribution shifts and novel scenarios. These limitations in robustness and generalization have become more apparent as autonomous vehicles (AVs) are expected to operate in real-world scenarios with complex, long-tail events that cannot be fully captured by traditional handcrafted rules \cite{chen2024end}. As a result, there is a growing recognition that the classical modular pipeline, while interpretable at an individual module level, lacks the flexibility and learning capability needed for truly autonomous driving in all conditions. Towards this end, end-to-end driving systems have emerged in recent years as a compelling alternative to jointly learn perception and control from raw sensor inputs by leveraging large-scale data.

 The overarching goal of end-to-end models is to optimize all tasks together that enables them outperform modular pipelines in challenging scenarios through joint feature learning. However, a typical end-to-end model often behaves as a ``black box,'' which raises concerns about its interpretability and reliability in autonomous driving. This development has guided the research direction towards foundation models and multimodal learning as a way of injecting structured knowledge and interpretability into typical end-to-end models \cite{gao2024survey, zhou2023vision+}. In particular, multimodal foundations that combine \textit{vision} and \textit{language} provide rich semantic understanding and reasoning. By unifying visual perception with high-level textual reasoning, a driving agent can potentially move beyond pure pattern recognition to a deeper semantic understanding of its surrounding environment.

Vision–Language Models (VLMs) have thus emerged as a promising framework in autonomous driving. These category of models are typically trained on paired image–text data to align visual content with natural language descriptions, thereby providing a driving agent with a human-like understanding of ``what is happening'' around it. Consequently, VLM-driven systems can perform tasks that were previously infeasible for vision-only systems. For instance, VLMs can address open-ended visual question answering (QA) about traffic situations and follow complex natural-language navigation instructions, all based on their internal understanding of the visual scene. Recent VLMs such as DriveGPT4 \cite{xu2024drivegpt4} highlight how vision-language systems can handle complex and multimodal tasks in driving. Rather than limiting output to fixed control signals, VLMs can explain what they see, describe what they intend to do, and follow language instructions. This means an AV could one day justify its actions in human-understandable terms (e.g., ``I am slowing down because I see a pedestrian crossing''), which leads to enhanced trust and safety.

Despite the enhanced semantic understanding and interpretability in autonomous driving systems by VLMs, they suffer from several limitations. One challenge is that most existing VLMs for autonomous driving can only cope with single-view inputs and lack efficient mechanisms to process multi-view sensor streams, which are standard in modern AVs. Some existing approaches attempt to integrate multi-view inputs through either feature concatenation or global attention across flattened representations, but such methods typically overlook the relative importance of individual views or feature channels\cite{zhou2024vision}. Another challenge is that many current VLMs adopt large-scale vision backbones and heavyweight language decoders\cite{shi2024we}. This makes them impractical for real-time deployment in embedded platforms with strict latency and memory constraints.

\begin{figure}
    \centering
    \includegraphics[width=\columnwidth]{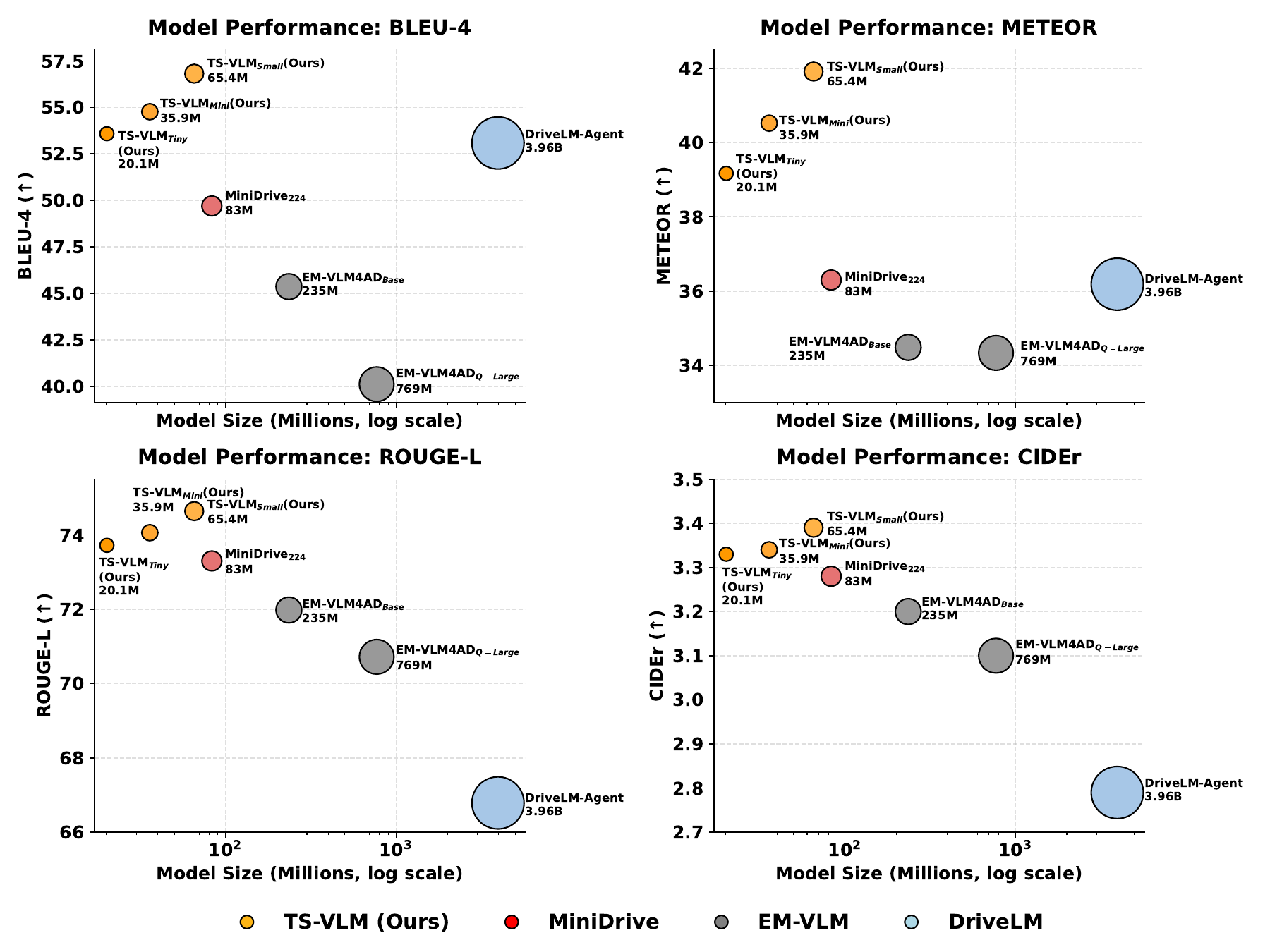}
    \caption{Model performance vs. model size on the DriveLM benchmark across four metrics: BLEU-4, METEOR, ROUGE-L, and CIDEr. Each circle represents a model, where the x-axis indicates model size (in millions of parameters, log scale), and the y-axis shows the performance score (↑ = better).}
    \label{fig:circles}
\end{figure}
In this work, we advance the state-of-the-art by introducing Text-Guided SoftSort Pooling VLM (TS-VLM), a novel lightweight VLM tailored for AVs. The proposed TS-VLM is designed to overcome the aforementioned challenges through a mobile-optimized vision backbone and a novel multi-view feature fusion strategy. The proposed vision encoder along with the adapter enables efficient perception and reasoning across multiple camera streams. In particular, TS-VLM leverages \textit{MobileViTv2} \cite{mehta2022separable} to drastically reduce the parameter count and computation per image, without sacrificing representational power. We further propose a Text-Guided SoftSort Pooling (TGSSP) that effectively aggregates information from multiple camera views, addressing the multi-view perception inefficiency in prior VLMs. In this respect, we evaluated the proposed TS-VLM on the DriveLM benchmark\cite{sima2023drivelm}, which achieved the state-of-the-art performance for visual QA tasks, while operating with considerably lower computational overhead (see Fig.\ref{fig:circles}). Towards this end, the contributions of this paper can be summarized as follows:

\begin{itemize}

    \item {We propose TS-VLM, with a considerably lower computational overhead compared to state-of-the-art models, which achieves superior accuracy and is suitable for real-time performance;}
    
    \item {We introduce TGSSP, a novel module to effectively fuse multi-view image features guided by input queries. This module enhances cross-modal alignment and improves the model's ability to interpret complex driving scenarios;}
    
    \item {We utilize MobileViTv2 as our visual encoder backbone to efficiently extract visual features. Its lightweight design contributes significantly to the overall reduction in parameters and computational cost, making real-time inference feasible on edge devices;}

    \item {We conduct thorough experiments on the DriveLM dataset to validate our approach. Our results demonstrate that TS-VLM achieves a BLEU-4 score of 56.82, METEOR of 41.91, ROUGE-L of 74.64, and CIDEr of 3.39, outperforming state-of-the-art. Our smallest variant contains only 20.1 M parameters and requires just 0.11 GB of memory during inference.}
\end{itemize}

The rest of the paper is organized as follows. In Section \ref{sec2}, we review recent advancements in VLMs and their integration with large language models (LLMs). We also explore the applications of these integrated models in the field of autonomous driving. The description of the DriveLM dataset along with the architecture of TS-VLM is detailed in Section \ref{sec3}. We present the attained results in Section \ref{sec4}, and Section \ref{sec5} concludes the paper. 

\section{Related Work}\label{sec2}
This section is devoted to reviewing the most recent VLMs and their integration with LLMs along with their applications in autonomous driving.

\subsection{Vision-Language Models}
In recent years, VLMs have rapidly advanced due to the availability of web-scale image-text data and improvements in model architectures. Unlike traditional vision models that require task-specific labeled data, VLMs learn cross-modal representations from massive image–text pairs harvested from the internet, enabling them to generalize zero-shot to new tasks. \textit{Dual-encoder} models such as CLIP\cite{radford2021learning} employ separate image and text encoders trained with a contrastive loss to align visual and linguistic features in a shared embedding space. By training on hundreds of millions of image-caption pairs, CLIP achieves impressive zero-shot performance on image classification and retrieval, and can effectively associate image content with open-ended text descriptions. However, dual-encoder models only enforce coarse alignment at the global embedding level. They lack fine-grained cross-modal interaction, which often leads to failures on tasks requiring detailed grounding or visual reasoning.

In contrast, \textit{fusion-based models} perform joint encoding of images and text via cross-attention, which allows token-level interactions between modalities. For instance, models in the VisualBERT\cite{li2019visualbert} and VL-BERT\cite{su2019vl} family use a multi-modal Transformer to fuse visual and textual features. These fusion models are typically pre-trained with objectives such as image-text matching, masked language modeling, and region-level alignment to learn fine-grained associations. However, because a fusion model processes an image-text pair together through a heavy cross-modal encoder, it becomes a bottleneck for large-scale inference. Beyond standard Transformer-based fusion, Qian et al.~\cite{qian2022scene} proposed a scene graph refinement network that integrates structured object relations into the cross-modal encoding process to enhance semantic reasoning in visual question answering.

Another category of fusion architectures adopt \textit{generative pre-training} to endow VLMs with open-ended language output abilities. Wang et al.\cite{wang2021simvlm} proposed SimVLM, which demonstrated a single prefix language modeling objective on large-scale noisy alt-text data that suffice to learn powerful multimodal features. By training end-to-end on $1.8$ billion image-text pairs with a prefix-LM objective, SimVLM achieved state-of-the-art results on a spectrum of vision-language reasoning benchmarks. Later, Li et al.\cite{li2022blip} introduced BLIP, which is jointly trained on multiple objectives, including contrastive image-text alignment, binary image-text matching, and image-conditioned caption generation, by using a unified Transformer that can operate in either encoder or decoder mode. The cost is that such multi-objective frameworks must carefully balance losses to avoid one capability dominating, and they generally require large-scale training data and compute. To alleviate this, Kong et al.~\cite{kong2024multimodality} proposed a multimodality self-distillation strategy to accelerate inference of vision-language pretrained models.

Other researchers have explored integrating \textit{LLMs} with vision encoders to create powerful VLMs capable of more complex reasoning and dialogue. DeepMind’s Flamingo\cite{alayrac2022flamingo} is one such model that bridges a pretrained vision backbone with a pretrained language model, enabling in-context few-shot learning for multimodal tasks. Flamingo introduced novel cross-attention layers that feed sequences of interleaved image and text data into an LLM, allowing it to generate answers or descriptions conditioned on images. The result was a single VLM that, without task-specific fine-tuning, achieved new state-of-the-art few-shot performance on image captioning and visual question answering (VQA) by simply being prompted with a handful of examples. Similarly, BLIP-2\cite{li2023blip} connects a vision Transformer to a frozen language model using a learned Query Transformer, so-called Q-Former, which distills visual inputs into a small set of tokens that the LLM can understand. This two-stage design leverages powerful pre-trained LLMs such as OPT\cite{zhang2022opt} and Vicuna\cite{vicuna2023} for language generation, while learning a lightweight interface for images, resulting in a compute-efficient architecture. Towards this end, recent VLM architectures can be categorized by their training strategies: contrastive learning for representation alignment, image-to-text generative modeling for direct captioning abilities, and multi-task or dual objectives (e.g. FLAVA\cite{singh2022flava}, CoCa\cite{yu2022coca}) that combine contrastive and generative losses. These large-scale models have shown remarkable capabilities in vision-language reasoning tasks.

\subsection{LLM/VLM in Autonomous Driving}
Recent research has investigated various applications of VLMs to autonomous driving. Unlike deep learning-based perception models that rely purely on visual inputs, VLMs enable vehicles to comprehend and describe their surroundings using natural language. 

One critical application of VLMs in autonomous driving is open-world perception, i.e., the ability to recognize and describe novel objects or scenarios beyond fixed training datasets. Traditional object detectors in AVs rely on closed-set classification. However, real-world driving encounters unseen obstacles, unusual road signs, and rare vehicle types, thus a more flexible perception framework is needed. For example, OpenScene \cite{peng2023openscene} aligns $3D$ point cloud features with CLIP’s image-text embedding space to enable open-vocabulary $3D$ semantic segmentation without manual labels. By training a $3D$ network to predict point features that match CLIP image features, OpenScene can identify novel classes or attributes in point clouds via natural language queries. Similarly, CLIP2Scene \cite{chen2023clip2scene} transfers knowledge from $2D$ CLIP into a LiDAR segmentation model using cross-modal contrastive learning to achieve label-efficient $3D$ scene understanding. This approach achieved the first zero-shot $3D$ segmentation results on driving datasets and improved performance with minimal fine-tuning. CLIP has also been applied to other perception tasks, such as object detection, where Ke et al.~\cite{ke2024vldadaptor} proposed VLDadaptor to adapt detectors across domains via vision-language distillation. More broadly, VLMs enable referring expression tasks in driving. It means that given a description such as ``the blue car waiting at the crosswalk,'' the system can ground that language to a specific $2D/3D$ object in the scene. This extends to tracking, where models follow an object over time based on a linguistic query. VLMs have also been used for anomaly detection in dashcam videos.  AnomalyCLIP \cite{zhou2023anomalyclip}, for instance, uses CLIP features to detect unusual events such as accidents and erratic agents by matching video embeddings to textual descriptions of anomalies. Using language allows definition of arbitrary ``anomaly'' scenarios that the model can flag even if not seen during training. Moreover, by leveraging LLM reasoning, frameworks such as LLM-AD \cite{elhafsi2023semantic} can interpret contextual clues to decide if something in the scene is semantically off. 

Beyond open-world perception, LLMs and VLMs are being explored as high-level planners for driving, thanks to their strong reasoning abilities and world knowledge. A notable example is GPT-Driver \cite{mao2023gpt}, which reframes motion planning as a text generation problem. In GPT-Driver, sensor observations are encoded as textual tokens, and prompts are fed to a GPT-3.5 model which then ``writes'' a sequence of future waypoints and even a justification for its decisions. With a specialized prompting and fine-tuning strategy, the GPT-Driver planner can output precise trajectory coordinates along with natural language explanations for each maneuver. Other works integrate language guidance into the decision-making process by using LLMs to interpret navigational goals or traffic rules. In this respect, the LMDrive \cite{shao2024lmdrive} demonstrated an LLM-controlled vehicle that takes descriptions of the scene and issues driving commands in an iterative loop. Likewise, OmniDrive \cite{wang2024omnidrive} proposes a holistic agent that uses an LLM to reason over multi-modal inputs and output high-level driving decisions, effectively functioning as a central brain that calls perception or control sub-modules as needed. An immediate benefit of such an approach is improved interpretability. This is because the planner can output explanations, through which it is easier to verify and trust its decisions. However, a challenge remains in the ability to execute language-derived plans reliably. Therefore, some frameworks keep LLMs in an advisory role, which ensures safety if the output of the language model is imperfect.

A key challenge in autonomous driving is the integration of multiple camera views and other sensors for holistic scene understanding. Vision-language models offer a unique way to fuse multi-view information by using descriptive language as an intermediate representation. To address this, EM-VLM4AD \cite{gopalkrishnan2024multi} introduced an efficient multi-frame VLM tailored for autonomous driving. The model incorporates a Gated Pooling Attention(GPA) module to efficiently aggregate information from multiple camera views, which led to enabling real-time reasoning over driving scenarios. Building on this, MiniDrive \cite{zhang2024minidrive} introduced an alternative lightweight VLM using a novel Feature-Engineering Mixture of Experts (FE-MoE) module and Dynamic Instruction Adapter (DI-Adapter), achieving state-of-the-art results on the DriveLM-nuScenes dataset.

Moreover, a more recent application of VLMs in autonomous driving is generating natural language summaries of each camera view, and then reasoning across those summaries to get a holistic understanding. For instance, one approach could caption the front-camera image as ``a cyclist on the right and a stop sign ahead,'' and the side-camera as ``a parked car on the curb,'' then use an LLM to infer the global scene (``need to stop for the sign and watch the cyclist passing the intersection''). Some works explicitly retrieve or integrate Bird’s Eye View (BEV) representations using language. BEV-CLIP \cite{wei2024bev} presents a multi-modal retrieval method where a text query is embedded with an LLM and matched against a database of BEV feature maps. Using CLIP’s visual features for BEV images and a language model for text, BEV-CLIP can successfully find scenes that correspond to a given description, demonstrating cross-view understanding of complex traffic scenarios. This capability could aid in localization or in the curation of datasets. Additionally, multi-modal VLMs have been used to align camera and LiDAR data with textual context, effectively tying $2D$ and $3D$ views together. An example is CLIP2Scene \cite{chen2023clip2scene}, which constructs pixel–point–text triplets so that images and point clouds share a common semantic space. 

\section{Methodology}\label{sec3}
In this section, we provide a brief introduction to the DriveLM dataset. Then it is followed by a detailed description of the proposed TS-VLM architecture.

\begin{figure}
    \centering
    \includegraphics[width=\columnwidth]{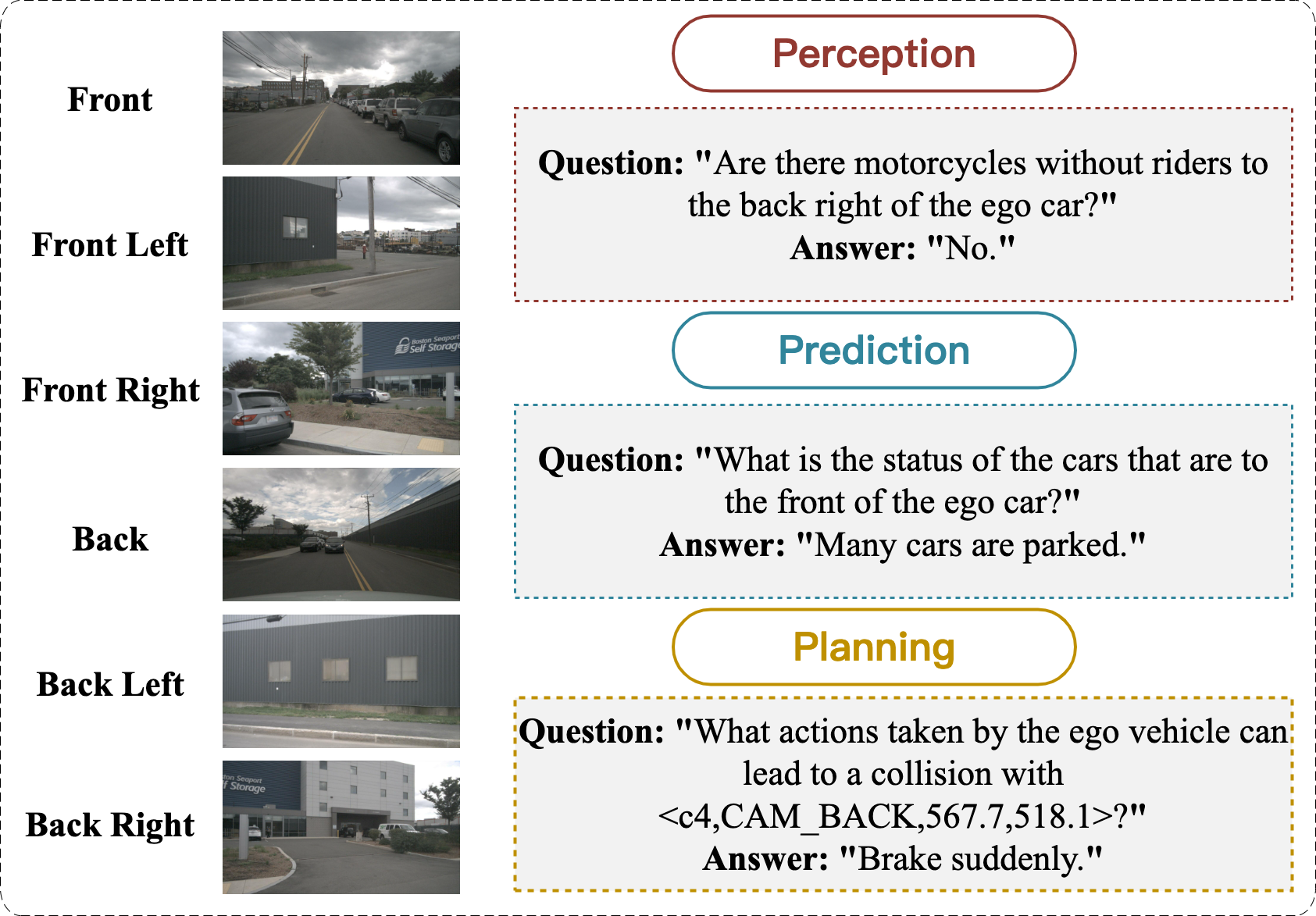}
    \caption{Examples of perception, prediction, and planning tasks in the DriveLM-nuScenes dataset.}
    \label{fig:DriveLM}
\end{figure}

\subsection{Dataset Description}
To evaluate the effectiveness of TS-VLM, we conduct experiments on the DriveLM-nuScenes benchmark \cite{sima2023drivelm}, which is a large-scale dataset designed to benchmark VLMs in autonomous driving. DriveLM-nuScenes provides multi-view images with corresponding natural language QA pairs, enabling the assessment of a model's ability to perform context-aware reasoning and decision-making in dynamic driving environments. This dataset includes $4,871$ frames and more than $440,000$ QA pairs. For each frame, there exist on average $91$ QA pairs, which are categorized into $3$ tasks including perception, prediction, and planning. Figures \ref{fig:DriveLM} illustrate examples of each task in this dataset. For training and evaluation, the dataset is divided into three subsets, where $90\%$ is used for training, $5\%$ for validation, and the remaining $5\%$ for testing.

\begin{figure}
    \centering
    \includegraphics[width=\columnwidth]{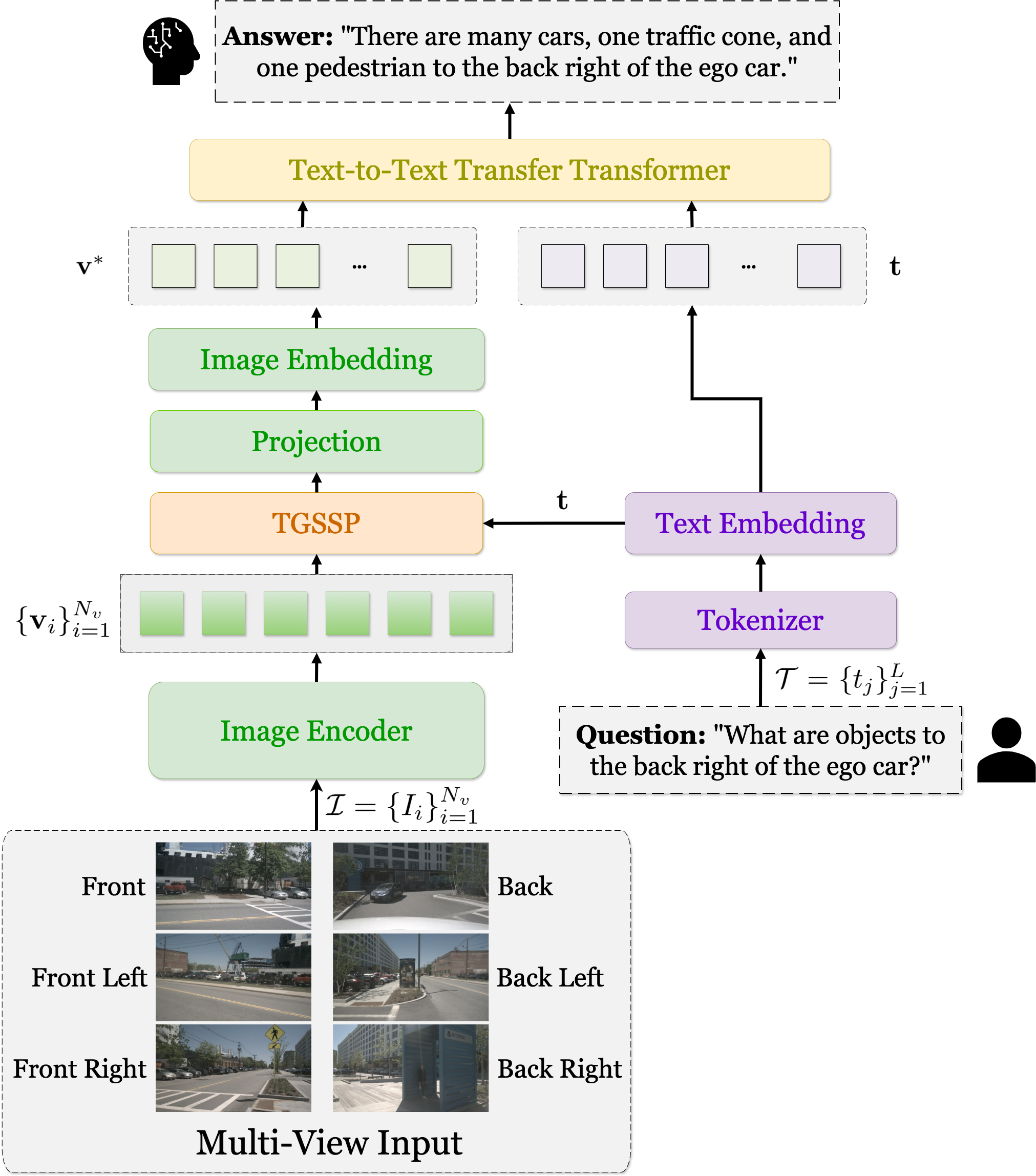}
    \caption{The overall architecture of TS-VLM. Multi-view images are first encoded via the image encoder and then adaptively fused with textual features using the Text-Guided SoftSort Pooling (TGSSP) module. The resulting multimodal embeddings are fed into the $T5$ LLM to generate semantically accurate textual answers based on the given question.}
    \label{fig:TS-VLM}
\end{figure}

\subsection{Model Architecture}
The overall architecture of the proposed model is illustrated in Fig. \ref{fig:TS-VLM}. Given multimodal input consisting of multiview images $\mathcal{I} = \{I_i\}_{i=1}^{N_v}$, and a textual query represented as tokens $\mathcal{T} = \{t_j\}_{j=1}^{L}$, the model aims to generate a textual response to accurately describe visual content. Here, \(N_v\) denotes the number of camera views and \(L\) is the length of the input text sequence.

Each image $I_i$ is first processed independently by the \textit{image encoder}, resulting in image embeddings $\{\mathbf{v}_i\}_{i=1}^{N_v}$, where $\mathbf{v}_i \in \mathbb{R}^{D_v}$ and \(D_v\) is the dimension of the visual embedding. In parallel, the textual query is converted into a textual embedding $\mathbf{t} \in \mathbb{R}^{L \times D_t}$ through tokenization and embedding operations, where \(D_t\) is the dimensionality of each text token representation. Subsequently, the visual embeddings $\{\mathbf{v}_i\}$ and the text embedding $\mathbf{t}$ are integrated by the TGSSP module, producing an aggregated visual embedding $\mathbf{v}^{*}$. Finally, the multimodal embedding that combines $\mathbf{v}^{*}$ and $\mathbf{t}$ is inputted to the LLM. In this work, we have used variants of the Text-to-Text Transfer Transformer (T$5$) variants (i.e., T$5$-small, T$5$-mini, and T$5$-tiny), to autoregressively generate the final textual output.

\begin{figure*}
    \centering
    \includegraphics[width=1\textwidth]{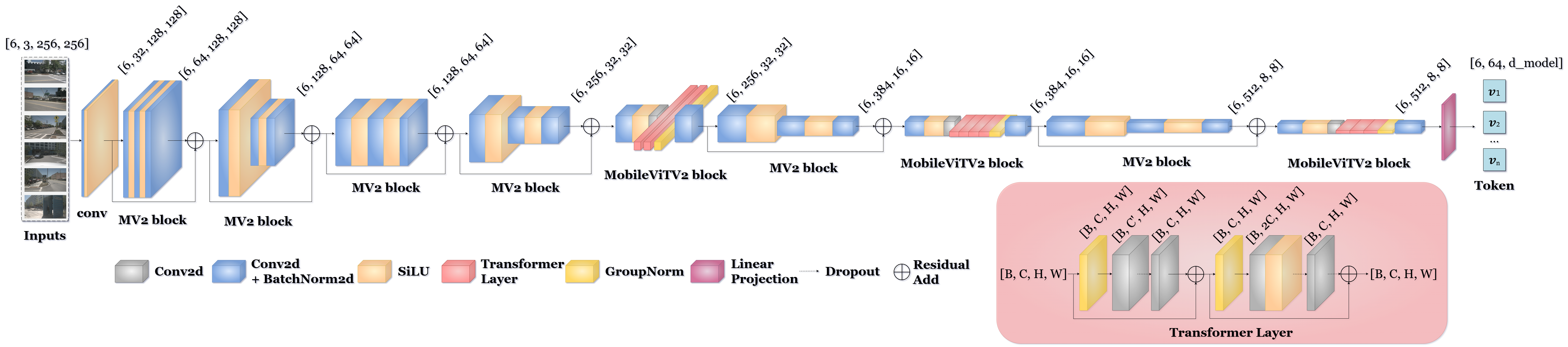}
    \caption{Structure of the proposed image encoder for TS-VLM.}
    \label{fig:Image Encoder}
\end{figure*}

\subsection{Image Encoder}
The overall structure of the proposed image encoder is illustrated in Fig. \ref{fig:Image Encoder}. As can be seen in this figure, the backbone network of the proposed image encoder is based on the MobileViTv2 (MV$2$) \cite{mehta2022separable} architecture. In this respect, the encoding process starts by applying a convolutional layer to reduce the spatial resolution of input images. Subsequently, multiple MV$2$ blocks are employed to efficiently extract spatial features. Each MV$2$ block consists of a point-wise convolution ($1\times1$ convolution), depth-wise convolution ($3\times3$ convolution), and another point-wise convolution, accompanied by residual connections to enhance feature propagation.

To further capture global dependencies in the visual data, MobileViTv2 blocks are introduced. Each of these blocks first performs depth-wise and point-wise convolutions, followed by an unfolding operation that prepares feature maps for separable self-attention. Within the self-attention sub-module, long-range dependencies among visual elements are explicitly modeled, allowing for the integration of both local and global visual contexts. After attention, a feed-forward network refines the representations, and the feature maps are folded back to their original spatial configuration. The process concludes with an additional point-wise convolution layer. Finally, a global pooling operation aggregates spatial information to produce compact embeddings that are subsequently projected into the multimodal embedding space through a linear transformation.

\begin{figure}
    \centering
    \includegraphics[width=\columnwidth]{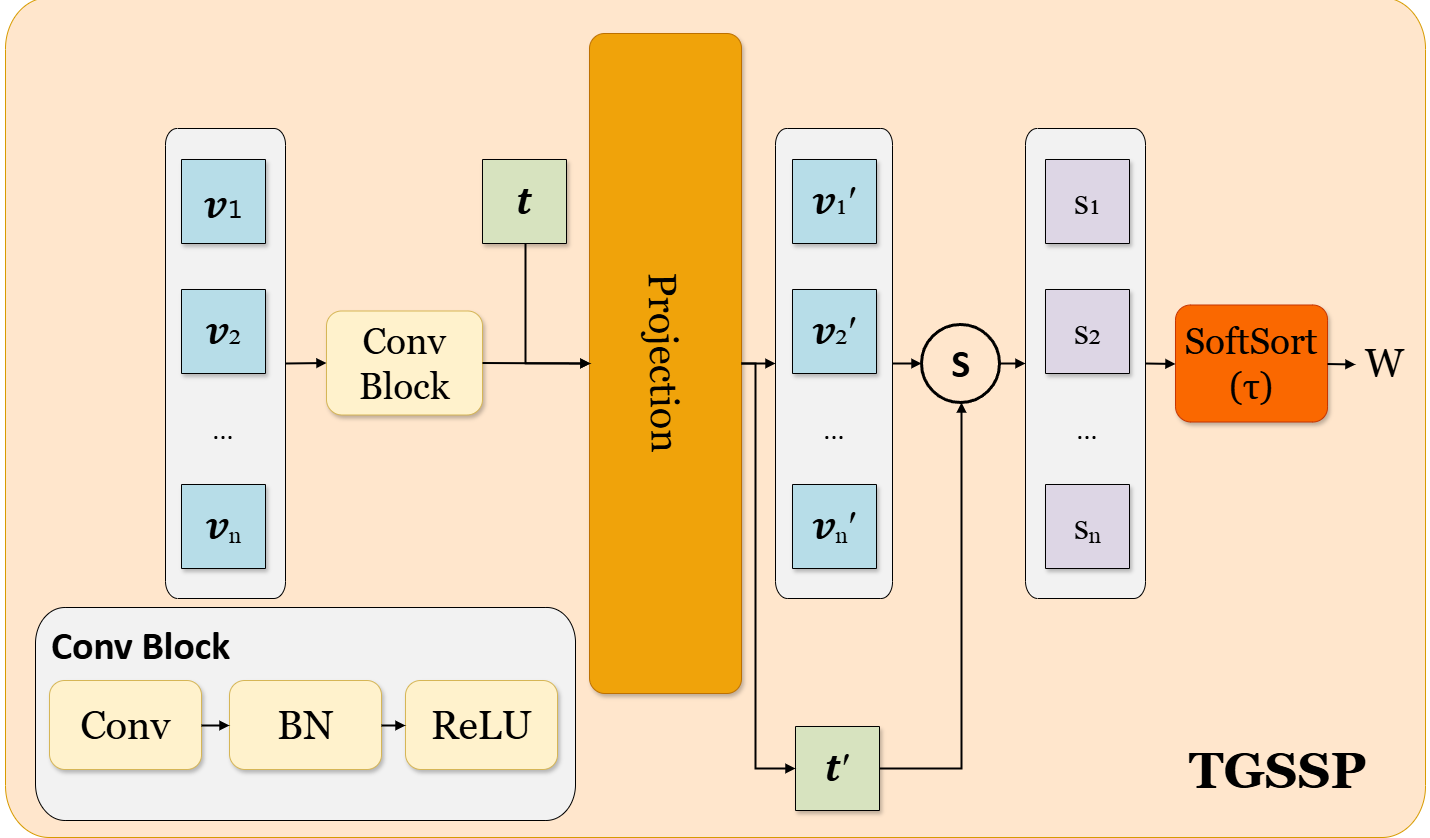}
    \caption{TGSSP Module.}
    \label{fig:TGSSP}
\end{figure}

\subsection{Text-Guided SoftSort Pooling (TGSSP)}
The overall architecture and data flow of the TGSSP module are detailed in Fig. \ref{fig:TGSSP}. The TGSSP module adaptively integrates visual information from multiple viewpoints guided by textual context. Given a set of visual embeddings $\{\mathbf{v}_i\}_{i=1}^{N_v}$, and a corresponding textual embedding \(t\) obtained from the token-level output of a pretrained T$5$ embedding layer, we first enhance the visual features via a convolutional block (Conv Block in Fig. \ref{fig:TGSSP}). This block comprises convolutional operations, batch normalization (BN), and a nonlinear activation (ReLU) to refine the visual features for improved representational quality. Subsequently, both enhanced visual embeddings and the original textual embedding are projected into a unified multimodal embedding space through a learned linear projection:
\begin{align}
  \mathbf{v}_i' & = \mathrm{Projection}(\mathrm{ConvBlock}(\mathbf{v}_i)),\\
\mathbf{t}' & = \mathrm{Projection}(\mathbf{t}).  
\end{align}

In this shared embedding space, the semantic similarity between each visual embedding and the textual embedding is explicitly computed to assess their alignment. Formally, the cosine similarity score \(s_i\) between each visual embedding \(\mathbf{v}_i'\) and the textual embedding \(\mathbf{t}'\) is computed:
\begin{align}
s_i = \mathrm{sim}(\mathbf{v}_i', \mathbf{t}') = \left\langle \frac{\mathbf{v}_i'}{\|\mathbf{v}_i'\|_2}, \frac{\mathbf{t}'}{\|\mathbf{t}'\|_2} \right\rangle,    
\end{align}
where $i = 1, 2, \dots, N_v$. These computed similarity scores are then fed into a differentiable sorting mechanism, SoftSort, characterized by a temperature scaling parameter \(\tau\). SoftSort smoothly approximates the sorting operation and generates a normalized weight vector \(W\) as follows:
\begin{align}
  \mathbf{W} = \mathrm{SoftSort}_\tau(s_1, s_2, \dots, s_{N_v}).  
\end{align}

These weights explicitly reflect the semantic relevance of each visual viewpoint with respect to the textual context, enabling a context-aware and selective visual feature aggregation. Thus, the TGSSP module dynamically assigns higher importance to visual embeddings that are semantically aligned closely with textual input.

\subsection{Language Model}
We employ three variants of the T$5$ including T$5$-small, T$5$-mini, and T$5$-tiny, as our language generation module. Given the multimodal embeddings from previous stages, these T$5$ models perform conditional text generation via an autoregressive decoding approach. Specifically, multimodal embeddings are treated as input tokens, and the T$5$ decoder sequentially generates output tokens by maximizing the conditional likelihood of the textual targets. 

\begin{table}
\centering
\setlength{\tabcolsep}{32pt}
\caption{Experimental Platform Configuration.}
\label{Configuration}
\begin{tabular}{lc}
\toprule
\textbf{Parameters} & \textbf{Configuration}\\ 
\cmidrule(lr){1-2}
CPU & Intel E5-2650 v4  \\ 
GPU & NVIDIA P100 \\ 
Memory & 32G \\ 
Architecture & Pytorch 2.3 + Cuda 11.8 \\ 
\bottomrule
\end{tabular}
\end{table}

\section{Experiments}\label{sec4}
In this section, we present the attained results by means of the TS-VLM for visual QA under the DriveLM benchmark. It is worth noting that all the experiments are performed under the same experimental environment, where its configuration is summarized in Table \ref{Configuration}.

\subsection{Evaluation Metrics}
To comprehensively evaluate the performance of TS-VLM, we employ a set of widely used natural language generation (NLG) metrics to measure the quality of generated textual outputs. Description of these metrics is as follows.

The \textit{Bilingual Evaluation Understudy (BLEU)} score \cite{papineni2002bleu} is a precision-based metric that measures $n$-gram overlap between the generated text and the reference text. It can be computed as below:
\begin{equation}
\text{BLEU-N} = \exp\left( \min\left(1 - \frac{l_r}{l_g}, 0\right) + \sum_{n=1}^{N} w_n \log p_n \right),
\end{equation}
where $l_g$ and $l_r$ are the lengths of the generated and reference sentences, respectively, $p_n$ represents the modified $n$-gram precision, and $w_n$ are weighting factors (typically uniform). A higher BLEU score indicates a better alignment of the generated answer with the reference text.

The \textit{Metric for Evaluation of Translation with Explicit ORdering (METEOR)} \cite{banerjee2005meteor} is an alternative to BLEU that incorporates synonym matching and stemming to improve semantic evaluation. Unlike BLEU, METEOR considers recall, precision, and alignment order using the following formula:
\begin{equation}
\text{METEOR} = F_{\text{mean}} \cdot (1 - \text{Penalty}),
\end{equation}
where $F_{\text{mean}}$ is the harmonic mean of precision and recall, and the \textit{penalty} term accounts for word order mismatches. METEOR has been shown to correlate better with human judgment than BLEU, which makes it particularly useful for evaluating model-generated textual descriptions in autonomous driving scenarios.

The \textit{Recall-Oriented Understudy for Gisting Evaluation (ROUGE-L)} \cite{lin2004rouge} measures the longest common subsequence (LCS) recall and precision between generated and reference texts as below:
\begin{equation}
\text{ROUGE-L} = \frac{LCS(G, R)}{|R|},
\end{equation}
where $LCS(G, R)$ is the longest common subsequence between the generated ($G$) and reference ($R$) sentences. ROUGE-L is particularly useful for assessing sequence-level similarity, making it well-suited for evaluating generated descriptions in VLM-based autonomous driving tasks.

Finally, the \textit{Consensus-based Image Description Evaluation (CIDEr)} metric \cite{vedantam2015cider} measures how well a generated sentence aligns with a set of reference captions. CIDEr uses the Term Frequency-Inverse Document Frequency (TF-IDF) weighting for $n$-grams and is computed as in the following:
\begin{equation}
\text{CIDEr} = \frac{1}{M} \sum_{m=1}^{M} \sum_{n=1}^{N} w_n \cdot \text{TF-IDF}(n),
\end{equation}
where $M$ is the number of reference sentences, $N$ is the maximum $n$-gram length, and $w_n$ is the TF-IDF weight for the $n$-gram. CIDEr emphasizes semantic relevance to ensure that the generated answers contain important driving-related details.

\begin{table*}
\centering
\setlength{\tabcolsep}{16pt}
\caption{Performance comparison of different models on the DriveLM-nuScenes benchmark (on the same test set).}
\label{tab:comparison}
    \begin{tabular}{lccccc} 
    \toprule
        \textbf{Model}&  \textbf{BLEU-4} ($\uparrow$) & \textbf{METEOR} ($\uparrow$) & \textbf{ROUGE-L} $(\uparrow$) & \textbf{CIDEr} ($\uparrow$)& \textbf{Parameters} ($\downarrow$)\\ 
        \cmidrule(lr){1-6}
         EM-VLM4AD$_{Base}$\cite{gopalkrishnan2024multi}&  45.36&  34.49&  71.98& 3.20& 235M\\ 
         EM-VLM4AD$_{Q-Large}$\cite{gopalkrishnan2024multi}&  40.11&  34.34&  70.72& 3.10& 769M\\ 
         DriveLM-Agent\cite{sima2023drivelm}&  53.09&  36.19&  66.79& 2.79& 3.96B\\ 
         MiniDrive$_{224}$\cite{zhang2024minidrive}&  49.70&  36.30&  73.30& 3.28& 83M\\ 
         MiniDrive$_{384}$\cite{zhang2024minidrive}&  50.20&  37.40&  73.50& 3.32& -\\ 
         TS-VLM$_{Small}$ (Ours)&  \textbf{56.82}&  \textbf{41.91}&  \textbf{74.64}& \textbf{3.39}& 65.4M\\ 
         TS-VLM$_{Mini}$ (Ours)&  \underline{54.77}&  \underline{40.52}&  \underline{74.06}& \underline{3.34}& \underline{35.9M}\\ 
         TS-VLM$_{Tiny}$ (Ours)&  53.59&  39.17&  73.72& 3.33& \textbf{20.1M}\\ 
         \bottomrule
    \end{tabular}
\end{table*}

\subsection{Results and Analysis}
Given the experimental setup and evaluation metrics, we compare TS-VLM with the current state-of-the-art VLMs in terms of evaluation metrics mentioned above, as well as the number of trainable parameters. Towards this end, the comparative results are presented in Table \ref{tab:comparison}. The attained results indicate that TS-VLM achieves the highest scores across all major evaluation metrics while maintaining a significantly lower parameter count than prior models. Compared to EM-VLM4AD and MiniDrive, TS-VLM demonstrates more effective multi-view feature fusion, leading to improved text generation quality. Notably, even the smallest variant, TS-VLM$_{Tiny}$, maintains competitive performance despite being significantly more compact, which verifies the efficiency of the proposed TGSSP module. While large-scale models such as DriveLM-Agent rely on extensive computational resources, TS-VLM surpasses them in key linguistic metrics despite being several orders of magnitude smaller. Such a trade-off is particularly critical for edge deployment scenarios.

\begin{table}
\centering
\caption{Computational analysis of different models.}
\label{tab:computational}
    \begin{tabular}{lccc} 
    \toprule
        \textbf{Model}&  \textbf{Parameters} &  \textbf{FLOPs}&  \textbf{Memory (GB)}\\ 
        \cmidrule(lr){1-4}
         EM-VLM4AD$_{Base}$\cite{gopalkrishnan2024multi}&  235M&  \underline{9.47B}&  0.94\\ 
         EM-VLM4AD$_{Q-Large}$\cite{gopalkrishnan2024multi}&  769M&  31.5B&  0.77\\ 
         DriveLM-Agent\cite{sima2023drivelm}&  3.96B&  439B&  14.43\\ 
         MiniDrive$_{224}$\cite{zhang2024minidrive}&  83M&  \textbf{5.9B}&  1.03\\
         DriveMLM\cite{wang2023drivemlm}&  8.37B&  535B&  36\\ 
         LLM-Driver\cite{chen2024driving}&  7B&  268B&  28\\ 
         Drive-GPT4\cite{xu2024drivegpt4}&  7.3B&  329B&  29.2\\ 
         TS-VLM$_{Small}$ (Ours)&  65.4M&  25.47B&  0.31\\ 
         TS-VLM$_{Mini}$ (Ours)&  \underline{35.9M}&  17.73B&  \underline{0.19}\\ 
         TS-VLM$_{Tiny}$ (Ours)&  \textbf{20.1M}&  13.81B&  \textbf{0.11}\\ 
         \bottomrule
    \end{tabular}
\end{table}

We further compare TS-VLM with existing models in terms of parameters,  Floating Point Operations (FLOPs), and memory usage. Results of this comparative analysis are summarized in Table~\ref{tab:computational}. Large models such as DriveMLM and Drive-GPT4 require significantly more computation and memory, making them impractical for real-time deployment. MiniDrive and EM-VLM4AD are more efficient but still have higher parameters and memory footprints compared to TS-VLM. Our model achieves the lowest resource consumption, with TS-VLM using only $20.1$M parameters and $0.11$GB memory, while maintaining strong performance. 

While TS-VLM presents moderately higher FLOPs compared to some small-scale baselines, the primary source of this cost lies in the TGSSP module. This operation involves a differentiable sorting layer that enables interpretable feature aggregation across multi-view inputs. Although this adds marginal computational overhead, it is both memory-efficient and highly parallelizable. As shown in Table~\ref{tab:inference}, TS-VLM variants maintain sub-$100$ms inference time per frame, confirming their suitability for real-time deployment in driving systems.

\begin{table}
\centering
\setlength{\tabcolsep}{20pt}
\caption{Inference time of TS-VLM variants.}
\label{tab:inference}
    \begin{tabular}{lc} 
    \toprule
        \textbf{Model} & \textbf{Inference Time (per frame)}\\ 
        \cmidrule(lr){1-2}
         TS-VLM$_{Small}$ (Ours)&  56.30 ms\\ 
         TS-VLM$_{Mini}$ (Ours)&  \underline{53.17 ms}\\ 
         TS-VLM$_{Tiny}$ (Ours)&  \textbf{49.59 ms}\\ 
         \bottomrule
    \end{tabular}
\end{table}

To systematically examine the impact of ranking strategies in our TGSSP module, we compare five additional alternatives, from structure-aware soft approximations to naïve averaging. All sorting experiments are conducted using our TS-VLM$_{Tiny}$ configuration, which serves as our most lightweight variant. 

SinkhornSort \cite{mena2018learning} uses an iterative approach to approximate permutation matrices. This is done by normalizing exponentiated similarity scores into a doubly stochastic form. While it introduces global alignment constraints, SinkhornSort leads to increased computational overhead. TopKSoft enforces sparsity by selecting the top-K most relevant views and applying softmax only within this subset, enabling focused yet differentiable attention. We set $K=3$ in our experiments to balance selectivity and information retention. SimpleSoftmax offers a minimal formulation by directly normalizing similarity scores without any positional modeling, serving as a lightweight but guided baseline. HardTop1 discards all but the most relevant view through hard masking. Finally, UniformPooling assigns equal weights to all views regardless of their relevance, providing a fully unguided baseline for fusion. The performance of these sorting algorithms are compared, where the quantitative results are summarized in Table~\ref{tab:sort_results}.

\begin{table*}
\centering
\setlength{\tabcolsep}{12pt}
\caption{Comparison of different sorting strategies used in our TGSSP module.}
\label{tab:sort_results}
    \begin{tabular}{lcccccccc} 
    \toprule
        \textbf{Sorting Method}&  \textbf{BLEU-1}& \textbf{BLEU-2} & \textbf{BLEU-3}& \textbf{BLEU-4} & \textbf{METEOR} & \textbf{ROUGE-L} & \textbf{CIDEr}& \textbf{FLOPs} \\ 
        \cmidrule(lr){1-9}
         SoftSort & \textbf{70.55} & \textbf{64.13} & \textbf{58.68} & \textbf{53.59} & \textbf{39.17} & \textbf{73.72} & \textbf{3.331} & 184 \\ 
         SinkhornSort & 69.90 & 63.56 & 58.19 & 53.14 & 38.72 & \underline{73.63} & \underline{3.321} & 33216 \\ 
         TopKSoft & \underline{70.51} & \underline{64.03} & \underline{58.52} & \underline{53.46} & \underline{39.00} & 73.48 & 3.308 & 208	\\ 
         SimpleSoftmax & 70.13 & 63.72 & 58.26 & 53.17 & 38.71 & 73.50 & 3.298 & 136 \\ 
         HardTop1 & 69.64 & 63.29 & 57.61 & 52.50 & 38.38 & 72.65 & 3.275 & 88 \\ 
         UniformPooling  & 68.11 & 61.58 & 55.91 & 50.80 & 37.16 & 72.22 & 3.249 & 48 \\ 
         \bottomrule
    \end{tabular}
\end{table*}

The attained results denote that SoftSort has the strongest and most stable results across all evaluation metrics. It offers a well-structured yet lightweight approach to ranking, achieving top scores while maintaining relatively low computational cost. This confirms its suitability as a default choice for TGSSP, which balances the attention precision and training stability. TopKSoft achieves nearly comparable results with SoftSort across several metrics, while introducing moderate sparsity into the ranking process. With FLOPs only marginally higher than SoftSort, it presents an appealing alternative in scenarios where sharper attention distributions are desired without compromising efficiency. SinkhornSort, however, fails to demonstrate significant advantages in output quality, despite its theoretical appeal as a structured permutation approximation. Its performance consistently lags behind SoftSort, while being more than 180 times more computationally expensive than SoftSort. This suggests that full permutation modeling is unnecessary in this context, and may even introduce unnecessary optimization complexity without benefiting final predictions. The two simpler strategies, i.e., SimpleSoftmax and HardTop1, further highlight the role of structure. SimpleSoftmax remains competitive despite the lack of explicit ordering due to its low complexity and stable training behavior, while HardTop1 scores drop significantly in metrics. Its extreme sparsity sacrifices contextual cues from secondary views, confirming that single-view pooling is insufficient for language-grounded multi-view reasoning. UniformPooling, which treats all views equally without guidance, performs the worst across all metrics. This baseline indicates the necessity of ranking mechanisms that, without relevance-weighted pooling, the fusion module fails to prioritize meaningful inputs and leads to semantically weaker outputs.

\begin{figure}
    \centering
    \includegraphics[width=\columnwidth]{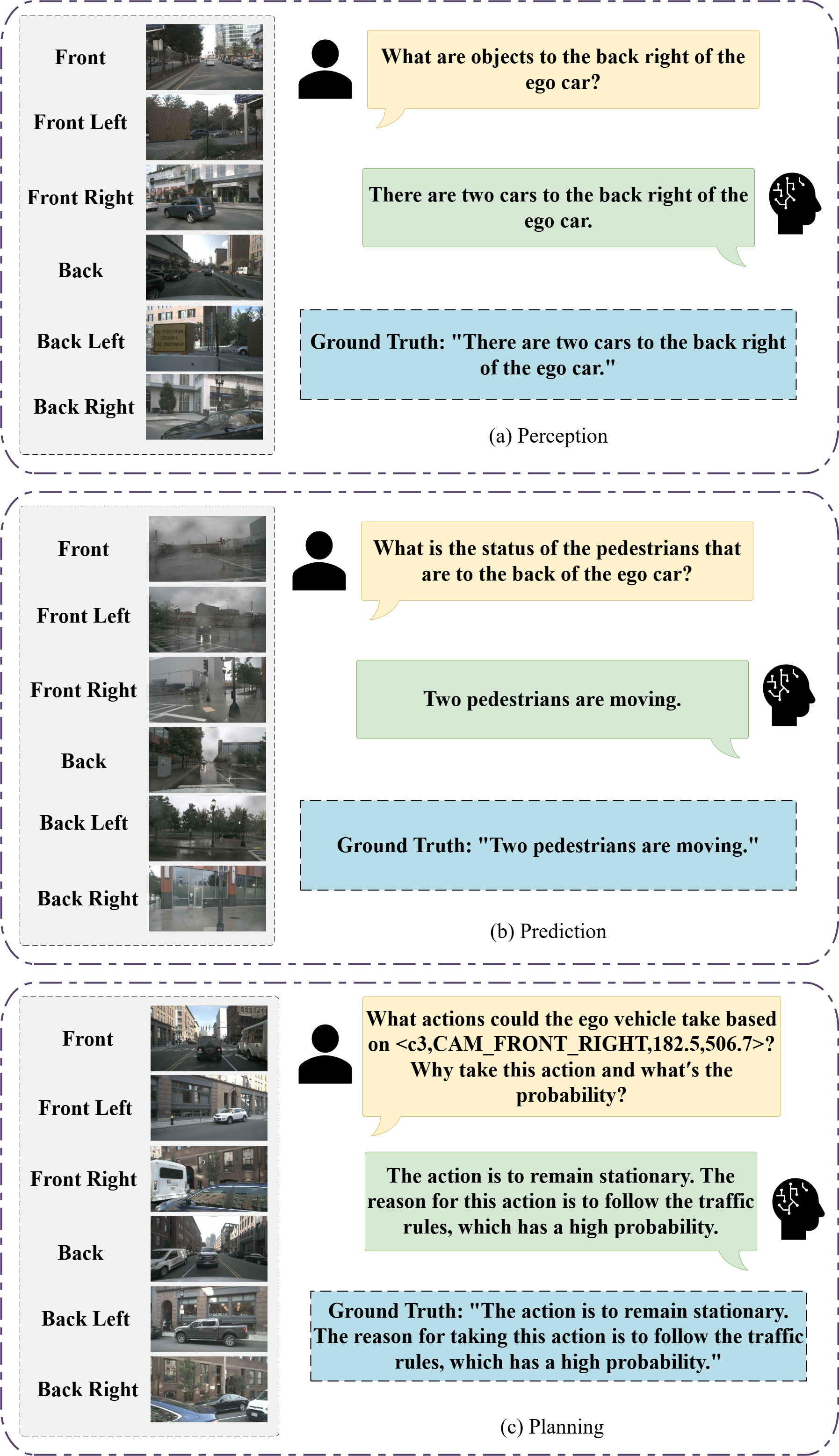}
    \caption{Examples of response generated by TS-VLM.}
    \label{fig:Generations}
\end{figure}

To present qualitative results of TS-VLM, we resort to visual QA under perception, prediction, and planning tasks, where the generated answers by means of TS-VLM are illustrated in Fig.~\ref{fig:Generations}. In the case of perception (Fig.~\ref{fig:Generations}-(a)), TS-VLM gives a direct and correct categorical response as compared with the ground truth. As for the prediction (Fig.~\ref{fig:Generations}-(b)), TS-VLM correctly predicts the status of objects in the scene with precise spatial grounding across views. In terms of planning (Fig.~\ref{fig:Generations}-(c)), TS-VLM not only predicts an appropriate high-level action (``remain stationary''), but also generates a plausible explanation grounded in traffic norms. Across all examples, the generated outputs closely align with the ground truth, which provides further evidence of TS-VLM’s effectiveness in complex, language-driven driving scenarios.

\subsection{Discussion}
TS-VLM performs well under tight memory and parameter constraints, but some components could be made more efficient. The SoftSort module, for example, introduces extra FLOPs during ranking. While this cost remains acceptable in practice, we are experimenting with lighter-weight alternatives that may offer similar functionality with lower compute. In addition, we conducted an ablation study to compare alternative sorting strategies within the TGSSP module. Although several lightweight variants demonstrate competitive results, SoftSort consistently achieves the best balance between accuracy and efficiency, reaffirming our design choice. This analysis further validates the role of differentiable ranking in TGSSP, and highlights the sensitivity of multi-view fusion to the form of attention weighting. Moreover, TS-VLM works on single-frame inputs. It does not yet take advantage of temporal patterns in driving videos, where motion or causality often spans multiple frames. We plan to expand TGSSP with temporal context in mind, allowing the model to track objects or evolving scenes across time. Finally, real-time deployment in onboard systems may benefit from architecture compression and GPU/TPU-specific optimization. Tailoring the model to specific deployment environments remains an important next step.

\section{Conclusion}\label{sec5}
This paper presents TS-VLM, a lightweight and efficient VLM for multimodal understanding in autonomous driving, which achieves state-of-the-art performance while significantly reduces computational costs. Through TGSSP, our model effectively aggregates multi-view information to address the limitations of previous pooling and attention-based fusion mechanisms. Experimental results on the DriveLM-nuScenes benchmark show that TS-VLM outperforms existing models in BLEU-4, METEOR, ROUGE-L, CIDEr and computational resources. These findings underscore the importance of optimized multi-view fusion strategies and lightweight architectures in developing practical VLMs. Future work will explore extending TS-VLM to incorporate temporal reasoning across video sequences and further optimizing inference speed for embedded automotive hardware.




 
%
\bibliographystyle{IEEEtran}
\bibliography{reference}

\vfill

\end{document}